\documentclass{article}
\usepackage[utf8]{inputenc}
\usepackage{amsmath}
\usepackage{amsfonts}
\usepackage{tabularray}
\usepackage{graphicx}
\usepackage{subcaption}
\usepackage{tabularray}
\usepackage{wrapfig}

\usepackage{xcolor}

\usepackage[preprint]{corl_2023} 

\title{Symmetric Models for Visual Force Policy Learning}

\author{
  Colin Kohler \\
  Khoury College of Computer Sciences \\
  Northeastern University \\
  United States \\
  \texttt{kohler.c@northeastern.edu} \\
  \And
  Anuj Shrivatsav Srikanth \\
  College of Engineering \\
  Northeastern University \\
  United States \\
  \texttt{srikanth.anu@northeastern.edu} \\
  \And
  Eshan Arora \\
  College of Engineering \\
  Northeastern University \\
  United States \\
  \texttt{arora.es@northeastern.edu} \\
  \And
  Robert Platt \\
  Khoury College of Computer Sciences \\
  Northeastern University \\
  United States \\
  \texttt{r.platt@@northeastern.edu} \\
}
\begin{document}

\maketitle

\begin{abstract}
While it is generally acknowledged that force feedback is beneficial to robotic control, applications of policy learning to robotic manipulation
typically only leverage visual feedback. Recently, symmetric neural models have been used to significantly improve the sample efficiency and 
performance of policy learning across a variety of robotic manipulation domains. This paper explores an application of symmetric policy learning to 
visual-force problems. We present Symmetric Visual Force Learning (SVFL), a novel method for robotic control which leverages visual and force feedback.
We demonstrate that SVFL can significantly outperform state of the art baselines for visual force learning and report several interesting empirical
findings related to the utility of learning force feedback control policies in both general manipulation tasks and scenarios with low visual acuity.
\end{abstract}


\section{Introduction}
\label{sec:intro}

There are a variety of manipulation tasks where it is essential to use both vision and force feedback as part of the control policy. Peg insertion with tight 
tolerances, for example, is a task that is nearly impossible to solve without leveraging force feedback in some form. The classical approach is to use
an admittance controller with a remote center of compliance to help the peg slide into the hole~\cite{whitney1987historical}. However, this is a very limited 
use of force feedback and it seems like it should be possible to use force information in a more comprehensive way.  Nevertheless, after decades of research,
it is still not clear how to accomplish this. One of the core obstacles is the difficulty in simulating the complex force interactions that happen at the robot end
effector. These depend upon the complex mechanics of the robotic drive train -- harmonic drives or planetary gearheads that cannot be modeled with any accuracy.
While there have been major efforts in the past to circumvent this challenge with series elastic drives~\cite{pratt1995series} or direct 
drives~\cite{asada1983control}, each of these approaches comes with its own set of challenges. 

An obvious alternative approach is to leverage machine learning, i.e. model free reinforcement learning (RL), to obtain force feedback assisted policies. This
is in contrast to vision-only RL where the policy only takes visual feedback~\cite{levine2016end, wang2020policy}. In \emph{visual force} RL, there is the 
possibility to adapt control policies directly to the mechanical characteristics of the system as they exist in the physical world, without the need to model
those dynamics first. However, this assumes that we can run RL online directly in the physical world, something that is hard to do due to the poor sample 
efficiency of RL. RL is well known to require an enormous amount of data in order to learn even simple policies effectively. While visual force 
RL might, in principle, be able to learn effective policies, this sample inefficiency prevents us from learning policies directly on physical equipment.
In order to improve the sample efficiency of RL in visual force problems, one common approach is to learn a helpful latent representation during 
a pretraining phase~\citep{chen2022visuo, lee2020making, zheng2020lifelong, fazeli2019see}. This generally takes the form of self-supervised robot ``play'' in 
the domain of interest that must precede actual policy learning. Unfortunately, this is both cumbersome and brittle as the latent representation does not generalize
well outside the situations experienced during the play phase. This is especially prevalent in the visual force domain as the noisy nature of force sensors 
means there will be many force observations not experienced during pretraining leading to poor latent predictions during policy learning.

This paper develops an alternative approach to the problem of visual force learning based on exploiting domain symmetries using equivariant 
learning~\citep{cohen2016steerable}. Recently, symmetric neural networks have been shown to dramatically improve the sample efficiency of
RL in robotic manipulation domains~\citep{wang2022so,wang2022equivariant}. However, this work has focused exclusively on visual feedback and has
not yet been applied to visual force learning. Here, there are several open questions. Can symmetric neural models improve sample efficiency in  problems with
force feedback? What might the model architecture look like to accomplish that? On what sorts of manipulation tasks might this approach be most helpful? 
This paper makes three main contributions. First, we propose a novel method for visual force policy learning called Symmetric Visual Force Learning (SVFL) 
which exploits the underlying symmetry of manipulation tasks to improve sample efficiency and performance. Second, we empirically evaluate the importance
of force feedback assisted control across a variety of manipulation domains and find that force feedback is helpful for nearly all tasks, not just contact-rich
domains like peg insertion where we would expect it to be important. Finally, we explore the role of force-assisted policies in domains with low visual acuity and characterize the degree to which force models can compensate for poor visual information.

\section{Related Work}
\label{sec:related}

\noindent \textbf{Contact-Rich Manipulation.} Contact-rich manipulation tasks, i.e. peg insertion, screw fastening, edge following, etc., are 
well-studied areas in robotic manipulation due to their prevalence in manufacturing domains. These tasks often are solved by hand-engineered polices 
which utilize force feedback and very accurate state estimation~\citep{whitney1987historical}, resulting in policies that perform well in structured 
environments but do not generalize to the large range of real-world variability. More recent work has proposed the use of reinforcement learning to address
these variations~\citep{levine2016end, kalakrishnan2011learning, zhu2018reinforcement} by training neural network policies which combine vision and 
proprioception. However, while these methods have been shown to perform well across a variety of domains and task variations, they require a high level 
of visual acuity, such that the task is solvable solely using image observations. In practice, this means these methods are unsuitable for a large portion of 
contact-rich manipulation tasks which require a high degree of precision and often include visual obstructions. 

\noindent \textbf{Multimodal Learning.} A common approach to multimodal learning is to first learn a latent dynamics model which compactly represents
the high-dimensional observations and then use this model for model-based learning. This technique has recently been adapted for use in various robotics
domains to combine various types of heterogeneous data. \citet{li2019connecting} combine vision and haptic information using a GAN but do not utilize 
their latent representation for manipulation policies. \citet{fazeli2019see} first learn a physics model using both vision and force data and use this
model as input to a handcrafted policy to play a game of Jenga. Similarly, \citet{zheng2020lifelong} propose a model which learns a
cross-modal visual-tactile model for a series of tasks, reusing past knowledge to perform lifelong learning. However, similar to~\citep{li2019connecting}
they do not use this learned representation for either a hand-crafted policy or policy learning. Our work is most closely related to~
\citep{lee2020making, chen2022visuo} which we use as baselines in this work. \citet{lee2020making} combine vision, force, and proprioceptive data using a 
variational latent model learned from self-supervision and use this model to learn a policy for peg insertion. \citet{chen2022visuo} learn a multimodal 
latent heatmap using a cross-modal visual-tactile transformer (VTT) which distributes attention spatially. They show that by combining VTT with stochastic
latent actor critic (SLAC), they can learn policies that can solve a number of manipulation tasks. In comparison to these works, we propose a sample-efficient
deterministic multimodal representation that is learned end-to-end without the need for pretraining. This is achieved through the use of a fully equivariant 
model which exploits the symmetry inherent in the $SO(2)$ domain to improve sample efficiency. Furthermore, we remove the need for the heavily structured, 
dense reward functions used in these previous works.

\noindent \textbf{Equivariant Neural Networks.} Equivariant networks were first introduced as G-Convolutions~\citep{cohen2016group} and Steerable
CNNs~\citep{cohen2016steerable, weiler2019general, cesa2021program}. Since their inception they have been applied across varied datatypes including
images~\citep{weiler2019general}, spherical data~\citep{cohen2018spherical}, and point clouds~\citep{dym2020universality}. More recent work has expanded 
the use of equivariant networks to reinforcement learning~\citep{wang2022equivariant, wang2020policy, wang2022surprising} and robotics~
\citep{simeonov2022neural, zhu2022sample, huang2022equivariant, kim2023se}. Compared to these prior works which focus on a single data modality, this 
works studies the effectiveness of combining various heterogeneous datatypes while preserving the symmetry inherit in each of these data modalities. 

\section{Background}
\label{sec:background}
\textbf{Equivariant Neural Networks.} A function is equivariant if it respects the symmetries of its input and output spaces. Specifically, a function
$f : X \rightarrow Y$ is \textit{equivariant} with respect to a symmetry group $G$ if it commutes with all transformations $g \in G, f(\rho_x(g)x) = \rho_y(g) f(x)$,
where $\rho_x$ and $\rho_y$ are the \textit{representations} of the group $G$ that define how the group element $g \in G$ acts on $x \in X$ and $y \in Y$,
respectively. An equivariant function is a mathematical way of expressing that $f$ is symmetric with respect to $G$: if we evaluate $f$ for various 
transformed versions of the same input, we should obtain transformed versions of the same output. Although this symmetry can be learned \cite{dehmamy2021automatic},
in this work we require the symmetry group $G$ and representation $\rho_x$ to be known at design time. For example, in a convolutional model, this can be accomplished
by  tying the kernel weights together to satisfy $K(gy) = \rho_{out}(g) K(y) \rho_{in}(g)^{-1}$, where $\rho_{in}$ and $\rho_{out}$ denote the representation of the 
group operator at the input and output of the layer~\citep{cohen2019general}. End-to-end equivariant models can be constructed by combining equivariant convolutional 
layers and equivariant activation functions. In order to leverage symmetry in this way, it is common to transform the input so that standard group representations
work correctly, e.g., to transform an image to a top-down view so that image rotations correspond to object rotations. 

\noindent \textbf{Extrinsic Equivariance.} Often real-world problems contain symmetry corruptions such as oblique viewing angles and occlusions. This is 
particularly prevalent in robotics domains where the state of the world is rarely fully observable. In these domains we consider the symmetry to be 
\textit{latent} where we know that some symmetry is present in the problem but cannot easily express how that symmetry acts in the input space. We 
refer to this relationship as \textit{extrinsic equivariance}~\citep{wang2022surprising}, where the equivariant constraint in the equivariant network
enforces equivariance to out-of-distribution data. While extrinsic equivariance is not ideal, it does not necessarily increase error and has been shown 
to provide significant performance improvements in reinforcement learning~\citep{wang2022surprising}.

\section{Approach}

\subsection{Problem Statement}
\label{sec:problem}

We model the visual force control problem as a discrete time finite horizon Markov decision process (MDP), $\mathcal{M} = (S, A, T, R, \gamma)$, 
where states $s \in S$ encode visual, force, and proprioceptive data and actions $a \in A$ command small end effector displacements. This MDP transitions
at a frequency of $20$ \textit{Hz} and the commanded hand displacements are provided as positional inputs to a lower level Cartesian space admittance 
controller that runs at $500$\textit{Hz} with a fixed stiffness. The hand is constrained to point straight down at the table (along the $-z$ direction).

State is a tuple $s = (I,f,e) \in S$. $I \in \mathbb{R}^{4 \times h \times w}$ is a 4-channel RGB-D image captured from a fixed camera pointed at the 
workspace. $f = (f_{xy}, f_z, m_{xy}, m_z) \in \mathbb{R}^{T \times 6}$ is a $T \times 6$ time series of the last $T$ measurements from a six-axis wrist
force-torque sensor transformed into the robot base frame. $e = (e_\lambda, e_{xy}, e_z, e_\theta) \in \mathbb{R}^5$ is the configuration of the end 
effector where $e_\lambda \in E_\lambda$ is the hand open width, $(e_{xy}, e_z)$ are the Cartesian coordinates of the hand, and $e_\theta$ is the 
orientation of the hand about the $-z$ axis. Actions are represented by $a = (\lambda, \Delta p) \in A \subseteq \mathbb{R}^5$ where $\lambda \in \mathbb{R}$
is the desired gripper open width and $\Delta p  = (\Delta p_{xy}, \Delta p_z, \Delta p_\theta) \in \mathbb{R}^{4}$ is the desired delta pose of the 
gripper with respect to the current pose $p$. As we discuss in the next section, we assume that the problem is $O(2)$-symmetric in the sense that the 
transition and reward functions are invariant with respect to planar rotations and reflections, for an appropriate definition of the action of $O(2)$ 
on $S$ and $A$.

\subsection{$\mathbf{O(2)}$ Symmetries in Visual Force Domains}
\label{sec:symmetry}

\begin{wrapfigure}[15]{r}{0.4\textwidth}
    \vspace{-0.25in}
    \centering
    \includegraphics[width=0.4\textwidth]{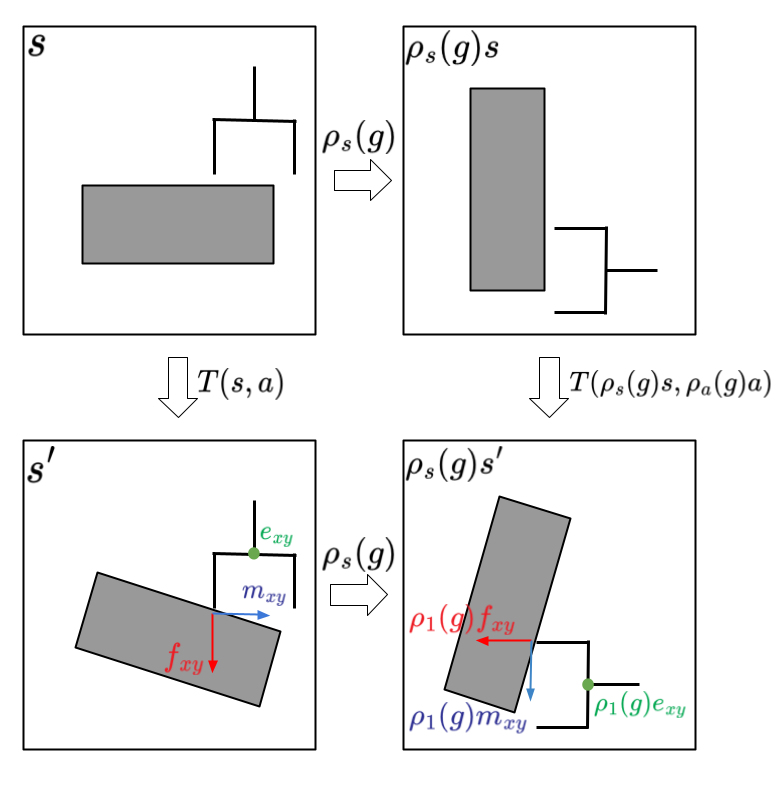}
    \caption{\textbf{$\mathbf{O(2)}$ Symmetries.}}
    \label{fig:symmetry}
\end{wrapfigure}

In order to leverage symmetric models for visual force policy learning, we utilize the group invariant MDP framework. A group invariant MDP 
is an MDP with reward and transition functions that are invariant under the group action, $R(s,a) = R(\rho_s(g) s, \rho_a(g)a)$ and 
$T(s,a,s') = T(\rho_s(g) s, \rho_a(g)a, \rho_s(g) s')$, for elements of an appropriate symmetry group $g \in G$~\cite{wang2022so}. $\rho_s$ 
and $\rho_a$ are representations of the group $G$ that define how group elements act on state and action. This paper focuses on
discrete subgroups of $O(2)$ such as the dihedral groups $D_4$ or $D_8$ that represent rotations and reflections in the $xy$ plane, i.e. the plane 
of the table. We utilize the $D_8$ group in our experiments (see Appendix \ref{sec:appendix:dn_size} for ablations on the effect of group size).

In order to express visual force manipulation as a group invariant MDP, we must define how the group operates on state and action such that the 
transition and reward invariance equalities described above are approximately satisfied. State is 
$s = (I,f,e) = (I, f_{xy}, f_z, m_{xy}, m_z, e_{xy}, e_z, e_\lambda)$. Since we are focused on rotations and reflections in the plane about the
$z$ axis, only the $xy$ variables are affected. Therefore, the group $g \in SO(2)$ acts on $s$ via 
$\rho_s(g) s = (\rho_0(g)I, \rho_1(g)f_{xy}, f_z, \rho_1(g)m_{xy}, m_z, \rho_1(g)e_{xy}, e_z, e_\lambda)$ where $\rho_0(g)$ is a linear operator 
that rotates/reflects the pixels in an image by $g$ and $\rho_1(g)$ is the standard representation of rotation/reflection in the form of a 
$2 \times 2$ orthonormal matrix. Turning to action, $a = (\lambda, \Delta p_{xy}, \Delta p_z, \Delta p_\theta)$, we define 
$\rho_a(g) a = (\lambda, \rho_1(g)\Delta p_{xy}, \Delta p_z, \Delta p_\theta)$. Given these definitions, visual force manipulation satisfies the
transition and reward invariance constraints, $R(s,a) = R(\rho_s(g) s, \rho_a(g) a)$ and $T(s,a,s') = T(\rho_s(g) s, \rho_a(g)a, \rho_s(g) s')$. 
This is illustrated for transition invariance in Figure~\ref{fig:symmetry}. 

\subsection{Model Architecture}
\label{sec:model}

\begin{wrapfigure}[11]{r}{0.4\textwidth}
    \vspace{-0.25in}
    \centering
    \includegraphics[width=0.4\textwidth]{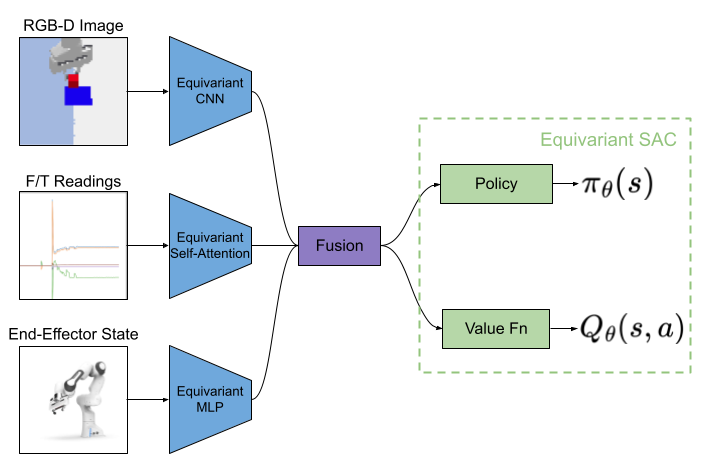}
    \caption{\small\textbf{High level model architecture.}}
    \label{fig:emf_model_diagram}
\end{wrapfigure}

As we discuss in the next section, we do policy learning using SAC which requires both a critic (a $Q$-function) and an actor. In our method,
both actor and critic employ the same encoder architecture which encodes state into a latent representation. Since our state $s = (I,f,e) \in S$ 
is multimodal (i.e. vision, force, and proprioception) our backbone is actually three encoders, the output of which is concatenated
(Figure~\ref{fig:emf_model_diagram}). The image encoder (top left in Figure~\ref{fig:emf_model_diagram}) is a series of seven equivariant convolutional 
layers. The force encoder (middle left) is a single equivariant self-attention layer. The proprioceptive encoder (bottom left) is a four-layer 
equivariant MLP. More details on each of these encoders can be found in the Appendix in Section~\ref{sec:appendix:network_arch}. In each of these 
encoders, the model respects the equivariance and invariance of each data modality corresponding to the relationships described in Section~\ref{sec:symmetry}. 

The force encoder is of particular note due to its use of single-headed self-attention. The input is a set of $T$
tokens, $f \in \mathbb{R}^{T \times 6}$, that encode the most recent $T$ measurements from the force-torque sensor. In order to make this model equivariant, 
we simply convert each of the key, query, and value networks to become equivariant models. For the standard implementation of self-attention, 
$\text{Attention} = \text{softmax}(f W^Q (f W^K)^T) f W^V$, the resulting group self attention operation is equivariant~\cite{romero2020group}:
\begin{align*} 
     \text{Attention}(X_f\Gamma) &= \text{softmax}(X_f \Gamma W^Q (X_f \Gamma W^K)^T) X_f \Gamma W^V   \\
                                 &= \text{softmax}(X_f W^Q \Gamma (X_f W^K \Gamma)^T) X_f W^V \Gamma   \\
                                 &= \text{softmax}(X_f W^Q \Gamma \Gamma^T (X_f W^K)^T) X_f W^V \Gamma \\
                                 &= \text{softmax}(X_f W^Q (X_f W^k)^T) X_f W^V \Gamma = \text{Attention}(X_f)\Gamma, 
\end{align*}
where, for simplicity of this analysis, we define $\Gamma$ to be the linear representation of the action of a group element $g \in G$ and 
$X_f \in \mathbb{R}^{T \times 6 \times |G|}$. \footnote{Although we omit the positional encoding here, this does not affect the result~\cite{romero2020group}.}
We informally explored alternative force-torque encoder models but found that this self attention approach worked best.

\subsection{Equivariant SAC}
\label{sec:equiv_sac}

For policy learning, we use Soft Actor Critic (SAC)~\citep{haarnoja2018soft} combined with the model architecture described above. This can be 
viewed as a variation of Equivariant SAC \citep{wang2022so} that is adapted to visual force control problems. The policy is a network 
$\pi : S \rightarrow A \times A_\sigma$, where $A_\sigma$ is the space of action standard deviations. We define the group action on the action
space of the policy network $\bar{a} \in A \times A_\sigma$ as: $\rho_{\bar{a}}(g) \bar{a} = (\rho_a(g) a, a_\sigma)$, where $a_\sigma \in A_\sigma$
and $g \in G$. The actor network $\pi$ is defined as a mapping $s \mapsto \bar{a}$ that satisfies the following equivariance constraint: 
$\pi(\rho_s(g)s) = \rho_a(g)(\pi(s))$. The critic is a Double Q-network: $q : S \times A \rightarrow \mathcal{R}$ that satisfies an invariant 
constraint: $q(\rho_s(g)s, \rho_a(g)a) = q(s,a)$.

\section{Experiments}
\label{sec:experiments}

We performed a series of experiments both in simulation and on physical hardware to validate our approach, Symmetric Visual Force Learning (SVFL). 
First, we benchmark SVFL's performance in simulation against alternative approaches in the literature. Second, we perform ablations that measure the contributions of 
different input modalities for different tasks under both ideal and degraded visual observations. Finally, we validate the approach on physical hardware.

\subsection{Simulated Experiments}
\label{sec:experiments:policy}

\begin{figure}
    \centering
    \includegraphics[width=0.90\textwidth]{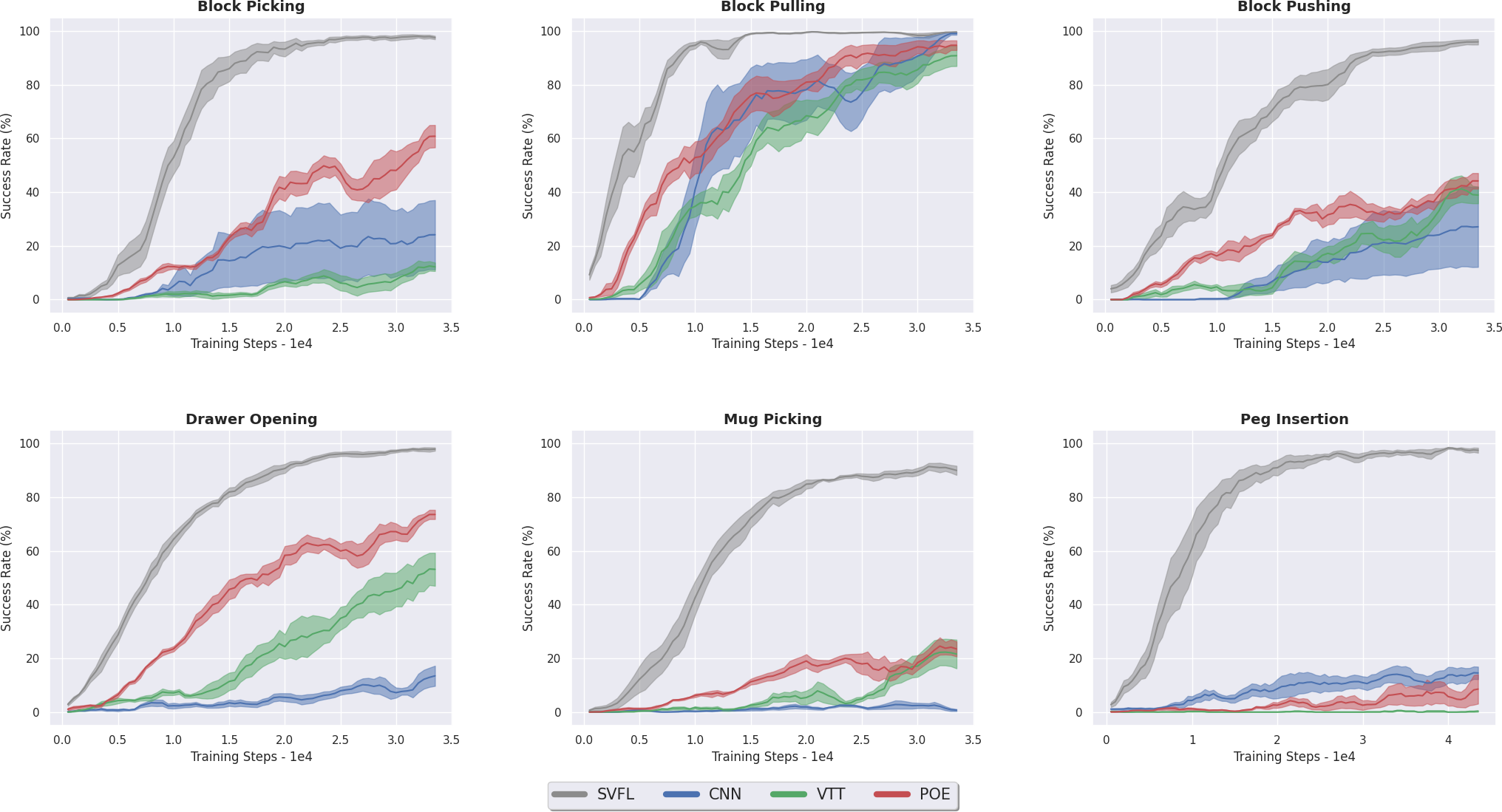}
    \caption{\textbf{Baseline Comparison.} Comparison of SVFL (gray) with baselines. Greedy evaluation policy is shown in terms of success rate. In all of our 
             experiments, results are averaged over 5 random seeds and the evaluation is performed every $500$ training steps. Shading denotes standard error.}
    \label{fig:baseline_comparison}
\end{figure}

\noindent \textbf{Tasks.} We evaluate SVFL across nine manipulation tasks from the BulletArm benchmark~\citep{wang2022bulletarm} which uses the
PyBullet~\citep{coumans2016pybullet} simulator: Block Picking, Block Pushing, Block Pulling, Block Corner Pulling, Mug Picking, Household Picking, 
Peg Insertion, Drawer Opening, and Drawer Closing (Figure \ref{fig:tasks}). For all tasks, a sparse reward function is used where a reward of $+1$ is given
at task completion and $0$ otherwise. Further task details can be found in the Appendix (Section \ref{sec:appendix:tasks}, \ref{sec:appendix:training_details}).

\textbf{Baselines.} We benchmark our method against two prominent alternative methods for visual force (or visual tactile) learning that have been proposed 
recently: Visual-Tactile Transformers (VTT)~\citep{chen2022visuo} and Product of Experts (PoE)~\citep{lee2020making}.
We also compare against a non-symmetric version of our model that is the same in every way except that it does not 
use equivariant layers (CNN). Both PoE and VTT are latent representation methods which rely on a self-supervised pretraining phase to build a compact latent
representation of the underlying states providing increased sample efficiency. Due to this pretraining, these methods represent attractive options for 
on-robot policy learning. While our method does not use any pretraining, and is therefore at a disadvantage relative to these two methods, we
maintained this pretraining phase as originally proposed in~\citep{lee2020making} and~\citep{chen2022visuo} as it is a core component of latent representation 
learning. In both baselines we used the encoder architectures proposed in~\cite{chen2022visuo} which were shown to outperform those in~\citep{lee2020making}.
PoE encodes the different input modalities independently using separate encoders and combines them using product of experts~\citep{lee2020making}. VTT
combines modalities by using self and cross-modal attention to build latent representations that focus attention on important task features in the visual
state space~\cite{chen2022visuo}. For further details on these baselines, see~\cite{chen2022visuo,lee2020making}. The latent encoders are pretrained for 
$10^4$ steps on expert data to predict the reconstruction of the state, contact and alignment embeddings, and the reward. All methods use Prioritized 
Experience Relay (PER)~\citep{schaul2015prioritized} pre-loaded with $50$ episodes of the expert data. For more details, see the Appendix 
(Section \ref{sec:appendix:training_details}). 

\noindent \textbf{Results.} We compared our method (SVFL) against the two baselines (POE and VTT) and the non-symmetric model (CNN) on the nine domains 
described above. Results from six representative domains are shown in Figure~\ref{fig:baseline_comparison} and results for all nine can be found in the
Appendix (Figure \ref{fig:full_baseline_comparison}). All results are averaged over five runs starting from independent random seeds. 
When compared to the baselines, SVFL has significantly higher success rates and sample efficiency in all cases.

\subsection{Sensor Modality Ablation} 
\label{sec:experiments:sensor_modality_study}

\begin{figure}
    \centering
    \includegraphics[width=0.90\textwidth]{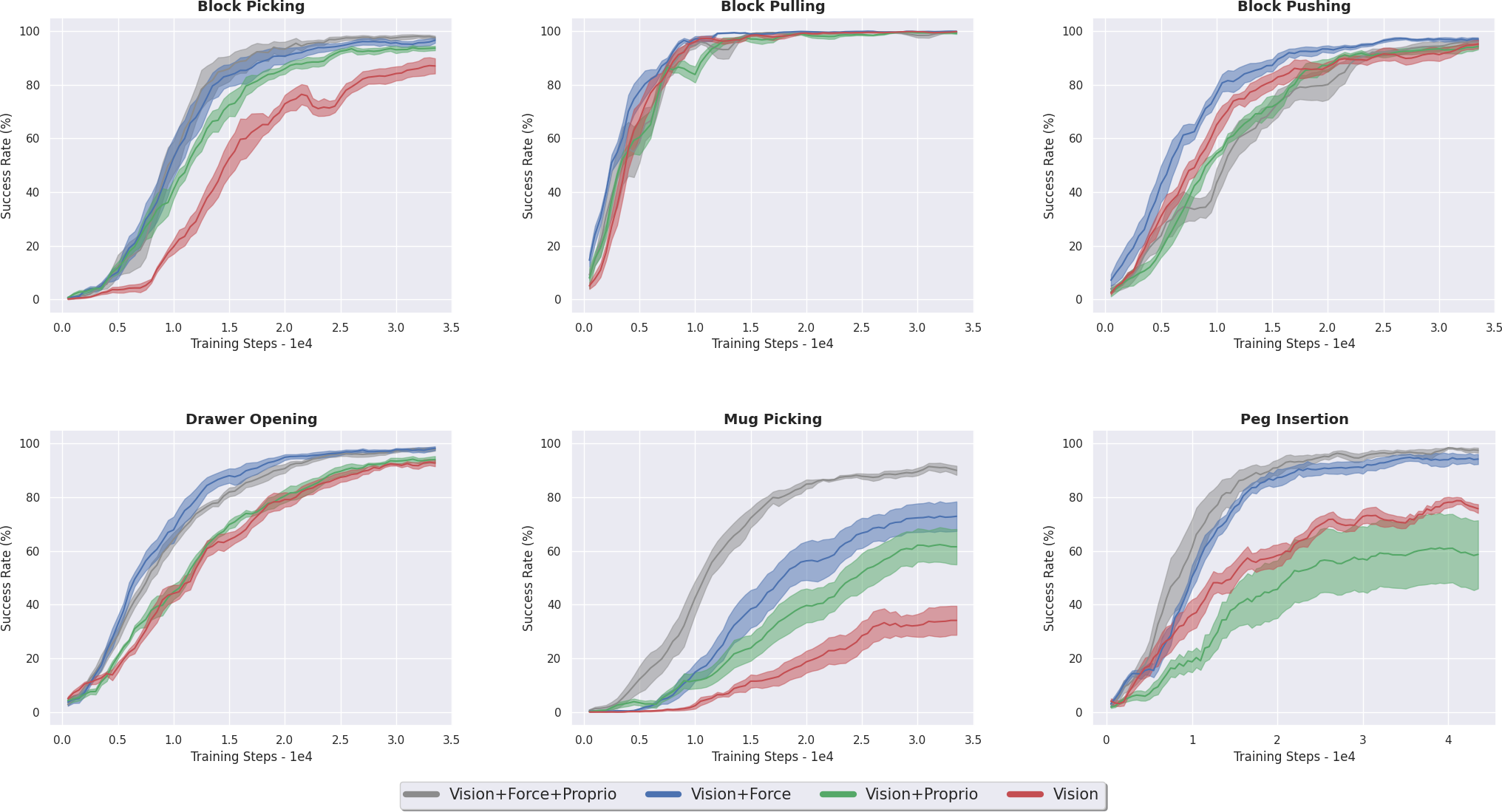}
    \caption{\textbf{Sensor Modality Ablation.} Comparison of the full SVFL model (gray) versus SVFL with subsets of the data modalities.}
    \label{fig:sensor_modality_study}
\end{figure}

Although it is intuitive that force data should help learn better policies on manipulation tasks, especially on contact rich tasks like peg 
insertion, it is important to validate this assumption and to measure the benefits that can be gained by using both vision and force feedback 
rather than vision alone. Recall that our state representation can be factored into three modalities, $s = (I,f,e)$, where $I$ is an image (vision),
$f$ is force, and $e$ is the configuration of the robot hand (proprioception). Here, we compare the performance of SVFL with all three modalities 
against a vision-only model, a vision/force model, and a vision/proprioception model on the same tasks as in Section \ref{sec:experiments:policy}.
Results for six tasks are shown in Figure~\ref{fig:sensor_modality_study} and complete results are given in Figure~\ref{fig:full_modality_ablation} 
in the Appendix. The results indicate that the inclusion of each additional sensor modality improves sample efficiency and performance 
for policy learning with all three sensor modalities performing best in most cases. However, notice that the degree to which force (and proprioceptive)
data helps depends upon the task. For example, the addition of force feedback drastically improves performance in Peg Insertion but has almost no 
effect in Block Pulling. There are, however, many tasks between these extremes. In Drawer Opening and Block Picking the force-aware policy converges
to a slightly higher success rate than the non-force assisted policies. The fact that force feedback is usually helpful, even in tasks where one might
not expect it, is interesting. This suggests that there is real value in incorporating force feedback into a robotic learning pipeline, even when there
is a non-trivial cost to doing so.

\subsection{Role of Force Feedback When Visual Acuity is Degraded}
\label{sec:experiments:reducded_visual_acuity}

\begin{figure}
    \centering
    \includegraphics[width=0.90\textwidth]{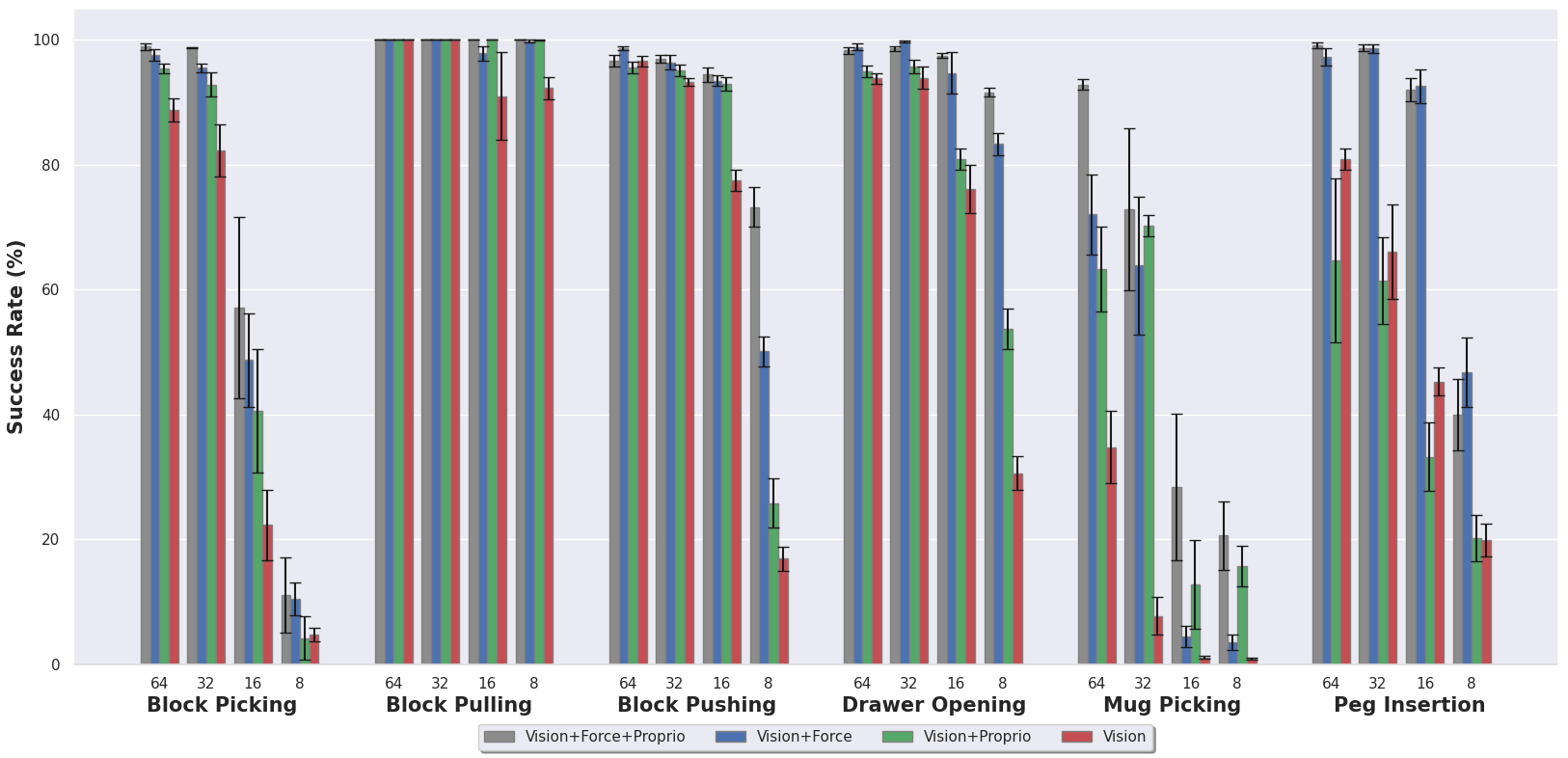}
    \caption{\textbf{Performance Under Degraded Visual Acuity.} Comparison of the full SVFL model (gray) versus SVFL with subsets of the data modalities 
             under visual acuity degradation. Performance is given after all models are trained to convergence.}
    \label{fig:reduced_img_exp}
\end{figure}

We also perform experiments in the context of degraded visual acuity to determine what happens if the visual input to our model is scaled down
significantly. Specifically, we evaluate the model on RGB-D images rescaled (bilinear interpolation) to four different sizes: $64 \times 64$, 
$32 \times 32$, $16 \times 16$, and $8 \times 8$. Aside from the rescaling, all other aspects of the model match the SVFL method detailed in 
the previous section. This experiment gives an indication of how force data can compensate for low resolution cameras, cloudy environments, or
smudged camera lenses. Figure~\ref{fig:reduced_img_exp} shows performance at convergence for six of the tasks at the four different levels of
visual resolution (see Figure~\ref{fig:full_img_res_ablation} in the supplementary material for corresponding results on all nine domains). 
We note several interesting observations. First, the importance of visual acuity is dependant on the task, e.g. high visual acuity is very important 
for Block Picking but not very important for Block Pulling. Second, force information generally tends to help the most in low visual acuity scenarios.
Finally, while force data generally improves performance, it cannot compensate for the loss of information in extreme visual degradation in tasks which
require high visual acuity.

\subsection{Real-World On-Robot Policy Learning}
\label{sec:experiments:real_world_eval}

\begin{figure}
    \centering
    \begin{subfigure}[c]{0.35\textwidth}
        \includegraphics[width=\textwidth]{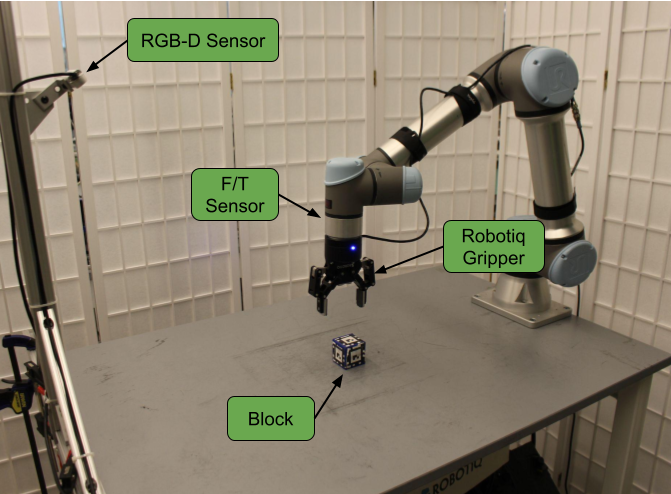}
    \end{subfigure}
    \begin{subfigure}[c]{0.44\textwidth}
        \includegraphics[width=\textwidth]{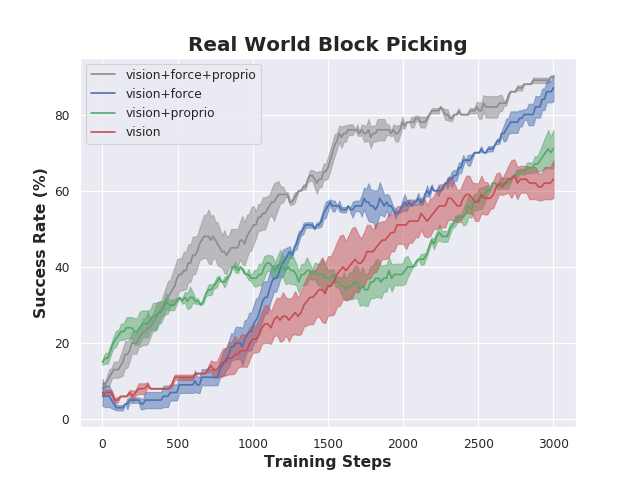}
    \end{subfigure}
    \caption{\textbf{On-Robot Policy Learning.} (Left) Robotic setup. (Right) Comparison of the full SVFL model (gray) versus SVFL with subsets of the data modalities
             in the real-world on the Block Picking task. Results are averaged over $3$ runs.}
    \label{fig:on_robot_learning}
\end{figure}

We repeat the simulated Block Picking policy learning experiment from Section \ref{sec:experiments:sensor_modality_study} in the real world to evaluate our methods
performance in the real-world. Figure~\ref{fig:on_robot_learning} shows the experimental setup which includes a UR5e robotic arm, a Robotiq Gripper, a wrist-mounted
force-torque sensor, and a Intel RealSense camera. The block is a $5mm$ wooden cube that is randomly posed in the workspace. We utilize AprilTags to track the block
for use in reward/termination checking and to automatically reset the workspace by moving the block to a new pose at the start of each episode. These tags are not
utilized during policy learning. In order to facilitate faster learning, we modify a number of environmental parameters in our real-world setup. We use a workspace
size as of $0.3m \times 0.3m \times \times 0.25m$ and a sparse reward function. We increase the number of expert demonstrations to $100$ (from $50$) and reduce the
maximum number of steps per episode to $25$ (from $50$). Additionally, we reduce the action space by removing control of the gripper orientation and increase the 
maximum amount of movement the policy can take in one step to $5cm$ (from $2.5cm$). We utilize the same model architecture as in Section \ref{sec:experiments:policy}.

Figure \ref{fig:on_robot_learning} shows the learning curve of the full SVFL model alongside the various subsets of data modalities available to our method. We train
all models for $3000$ steps taking around $4$ hours. As in the simulation results, the full SVFL model is both more sample efficient and outperforms SVFL. Additionally,
we see that force sensing is a vital component in this setting with the force-aware models achieving a $90\%$ success rate compared to the $60\%$ success rate of the 
non-force aware models (at $3000$ training steps).

\section{Discussion \& Limitations}
\label{sec:discussion}

This paper proposes Symmetric Visual Force Learning (SVFL), an approach to policy learning with visual force feedback that incorporates $SE(2)$ symmetries
into the model. Our experiments demonstrate that SVFL outperforms two recent high profile benchmarks, PoE~\citep{lee2020making} and VTT~\citep{chen2022visuo},
by a significant margin both in terms of learning speed and final performance. We also report a couple of interesting empirical findings. First, we find 
that force feedback is helpful across a wide variety of policy learning scenarios, not just those where one would expect force feedback to help, i.e. Peg 
Insertion. Second, we find that the positive effect of incorporating force feedback increases as visual acuity decreases. A limitation of this work is that
although we expect that our framework is extensible to haptic feedback, this paper focuses on force feedback only. Additionally, we constrain our problem
to top-down manipulation and planar symmetries in $SE(2)$ and therefor there is significant scope to extend this to $SE(3)$ symmetries. Finally, this paper
focuses primarily on RL but the encoder architectures should be widely applicable to other learning techniques such as imitation learning.

\bibliography{references}

\clearpage

\section{Appendix}
\label{sec:appendix}

\subsection{Manipulation Tasks}
\label{sec:appendix:tasks}

\begin{figure}[h!]
    \begin{subfigure}{0.24\textwidth}
        \includegraphics[width=\hsize]{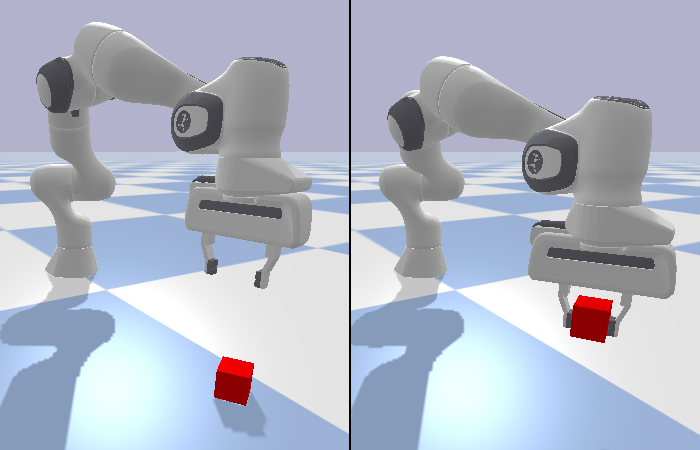}
        \caption{\tiny Block Picking}
        \label{fig:block_picking_ex}
    \end{subfigure}
    \begin{subfigure}{0.24\textwidth}
        \includegraphics[width=\hsize]{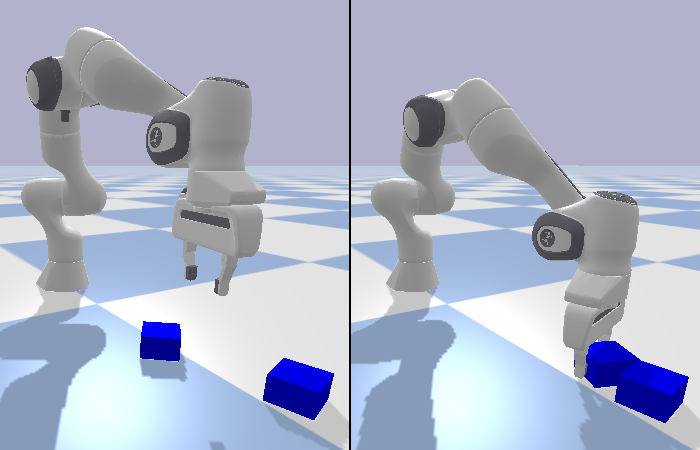}
        \caption{\tiny Block Pulling}
        \label{fig:block_pulling_ex}
    \end{subfigure}
    \begin{subfigure}{0.24\textwidth}
        \includegraphics[width=\hsize]{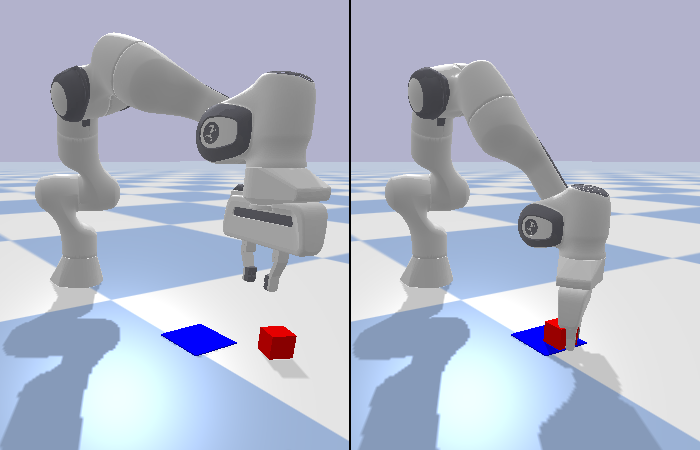}
        \caption{\tiny Block Pushing}
        \label{fig:block_pushing_ex}
    \end{subfigure}
    \begin{subfigure}{0.24\textwidth}
        \includegraphics[width=\hsize]{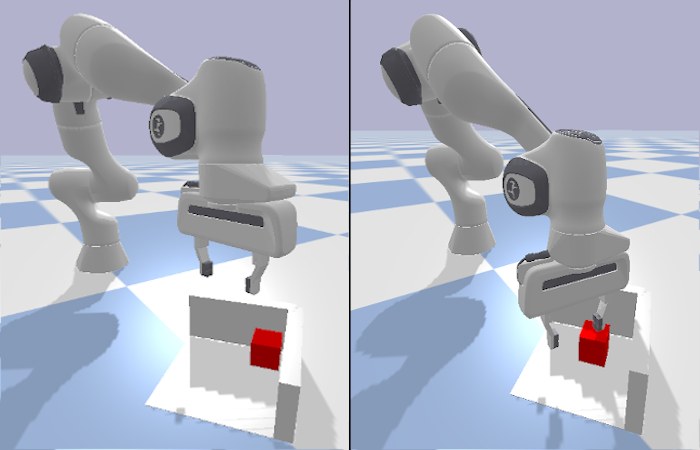}
        \caption{\tiny Block Pulling Corner}
        \label{fig:block_pulling_corner_ex}
    \end{subfigure} \\
    \begin{subfigure}{0.24\textwidth}
        \includegraphics[width=\hsize]{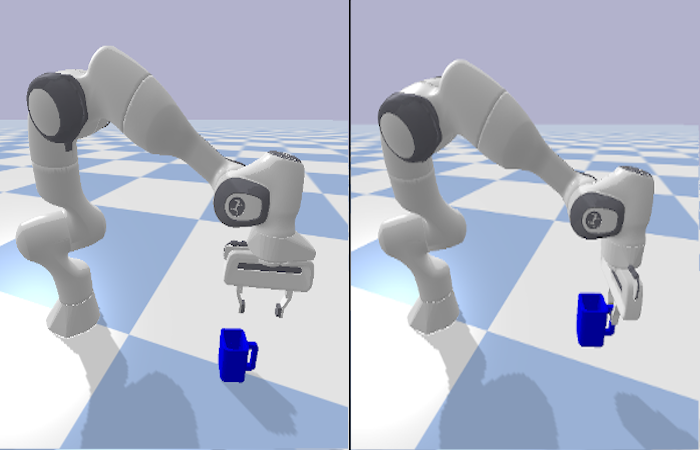}
        \caption{\tiny Mug Picking}
        \label{fig:mug_picking}
    \end{subfigure}
    \begin{subfigure}{0.24\textwidth}
        \includegraphics[width=\hsize]{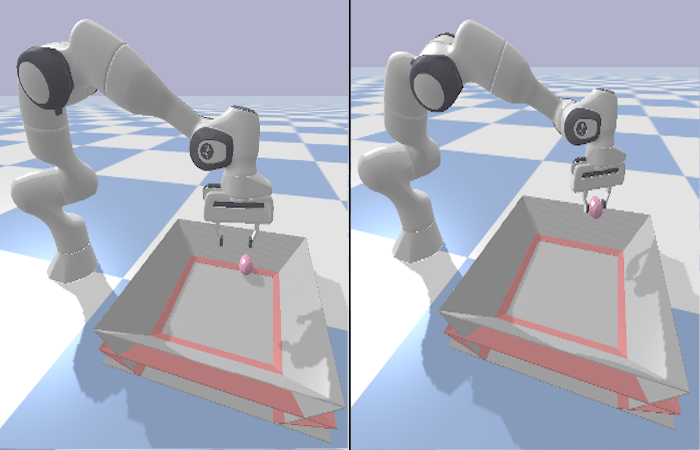}
        \caption{\tiny Household Picking}
        \label{fig:household_picking_ex}
    \end{subfigure}
    \begin{subfigure}{0.24\textwidth}
        \includegraphics[width=\hsize]{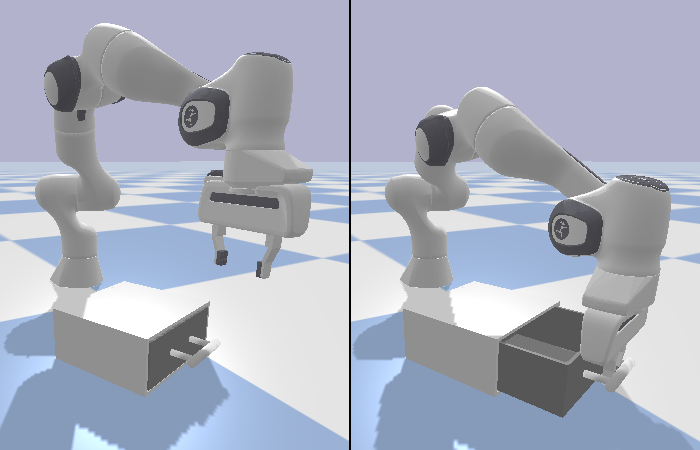}
        \caption{\tiny Drawer Opening}
        \label{fig:drawer_opening_ex}
    \end{subfigure} 
    \begin{subfigure}{0.24\textwidth}
        \includegraphics[width=\hsize]{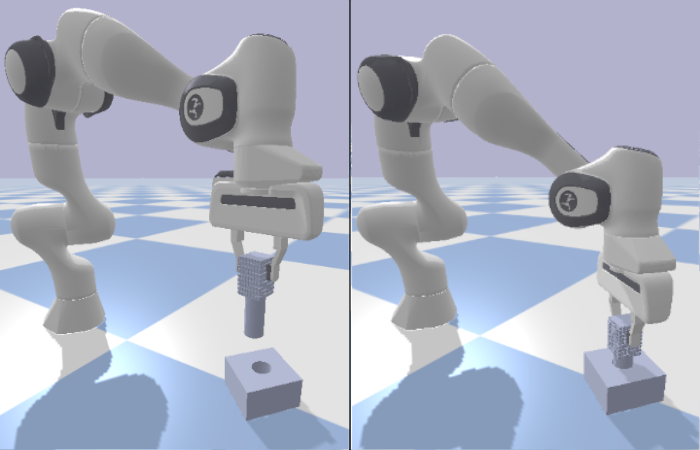}
        \caption{\tiny Peg Insertion}
        \label{fig:peg_insertion_ex}
    \end{subfigure}
    \caption{\textbf{Tasks.} The manipulation domains from the BulletArm benchmark \citep{wang2022bulletarm} implemented in PyBullet \citep{coumans2016pybullet}. 
             (Left) Initial state. (Right) Goal state.}
    \label{fig:tasks}
\end{figure}

We benchmark SVFL and the baselines across nine simulated tasks using the BulletArm Benchmark \cite{wang2022bulletarm} implemented in the PyBullet simulator
\cite{coumans2016pybullet}. The initial and goal states of each of these tasks can be seen in Figure \ref{fig:tasks}. All tasks use a sparse reward function 
where a reward of $+1$ is returned at task completion and $0$ otherwise. Task definition and parameters are detailed below. Further details about each of
these tasks can be found in the BulletArm benchmark \citep{wang2022bulletarm}. 

\noindent \textbf{Block Picking:} Pick up a cubic block and lift it to a specified height. In this task, we vary the initial pose, mass, size, and friction
parameters of the block.

\noindent \textbf{Block Pulling:} Pull two cubic blocks together so that they are touching. In this task, we vary the initial poses, masses, sizes, and friction
parameters of both blocks.

\noindent \textbf{Block Pushing:} Push a cubic block to a target position indicated by a blue marker. In this task, we vary the initial pose, mass, size, and friction
parameters of the block.

\noindent \textbf{Block Pulling Corner:} Pull a cubic block away from its initial pose nestled in the corner of a fixture. Due to the positioning of the block 
against these two walls, the robot must drag the block away from the corner using the tips of its gripper. In this task, we vary the initial pose, mass, size,
and friction parameters of the block and the initial pose and fiction parameters of the fixture. 

\noindent \textbf{Mug Picking:} Grasp a mug by its handle and lift it to a specified height. Grasping the mug in any other manner is not considered a success.
In this task, we vary the initial pose, mass, and size of the mug.

\noindent \textbf{Household Picking:} Grasp a randomly selected household object and lift it to a specified height. At the start of each episode, a random common 
household object is placed in a bin from a large collection of such objects. In this task, the object and its initial pose are randomized.

\noindent \textbf{Drawer Opening/Closing:} In drawer opening, the robot is tasked with pulling a drawer open using its handle until the drawer is opened to a 
specified position. Similarly, in drawer closing the robot is tasked with closing a drawer which is initialized to the open configuration. In both of these tasks, 
the initial pose of the drawer is randomized.

\noindent \textbf{Peg Insertion:} Insert a round peg into a round hole. The peg is modified with a square handle to provide the robot with a more stable grip on 
the peg and the task is initialized with the robot gripping the peg. In this task, only the initial pose of the hole is randomized.

\subsection{Network Architectures}
\label{sec:appendix:network_arch}

\begin{figure}
    \hspace{3.0em}
    \begin{subfigure}[t]{0.75\textwidth}
        \includegraphics[width=\hsize]{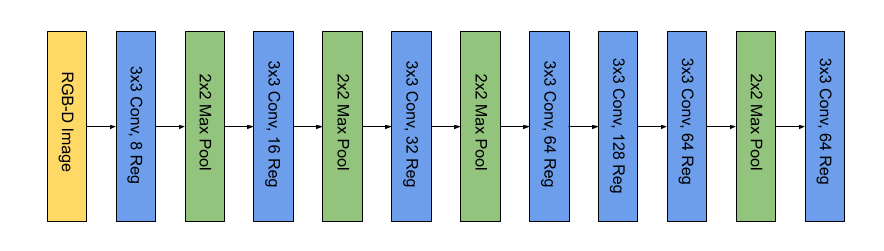}
        \caption{\textbf{Vision Architecture}}
        \label{fig:vision_network_diagram}
    \end{subfigure} \\
    \begin{subfigure}[t]{0.45\textwidth}
        \hspace{3em}
        \includegraphics[width=\hsize]{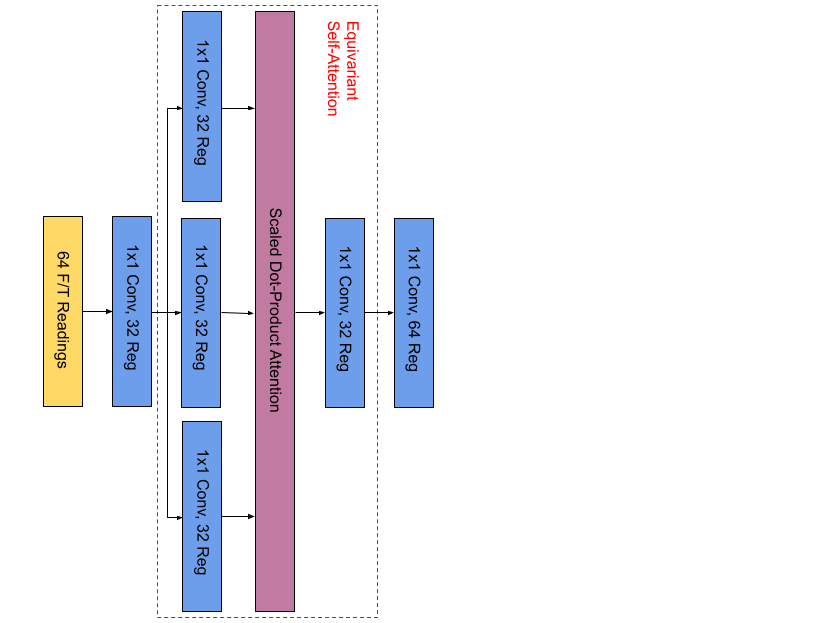}
        \caption{\textbf{Force Architecture}}Symmetric Models for Visual Force Policy Learning
        \label{fig:force_network_diagram}
    \end{subfigure}
    \begin{subfigure}[t]{0.45\textwidth}
        \includegraphics[width=\hsize]{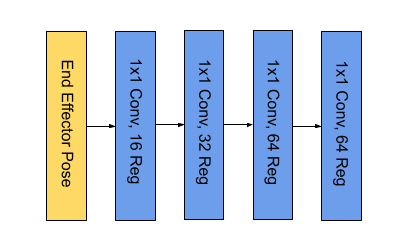}
        \caption{\textbf{Proprioception Architecture}}
        \label{fig:proprio_network_diagram}
    \end{subfigure}
    \caption{\textbf{Equivariant Encoder Architectures.} Network architectures of the equivariant encoders used in SVFL.}
    \label{fig:encoder_model_architectures}
\end{figure}

\begin{figure}
    \begin{subfigure}[t]{0.45\textwidth}
        \includegraphics[width=\hsize]{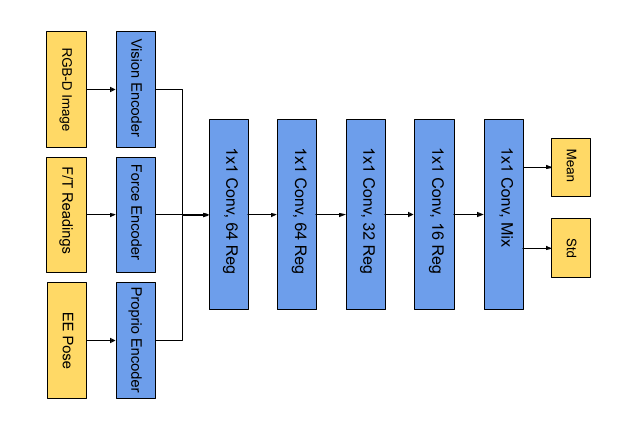}
        \caption{\textbf{Equivariant Actor Architecture.}}
    \end{subfigure}
    \begin{subfigure}[t]{0.45\textwidth}
        \includegraphics[width=\hsize]{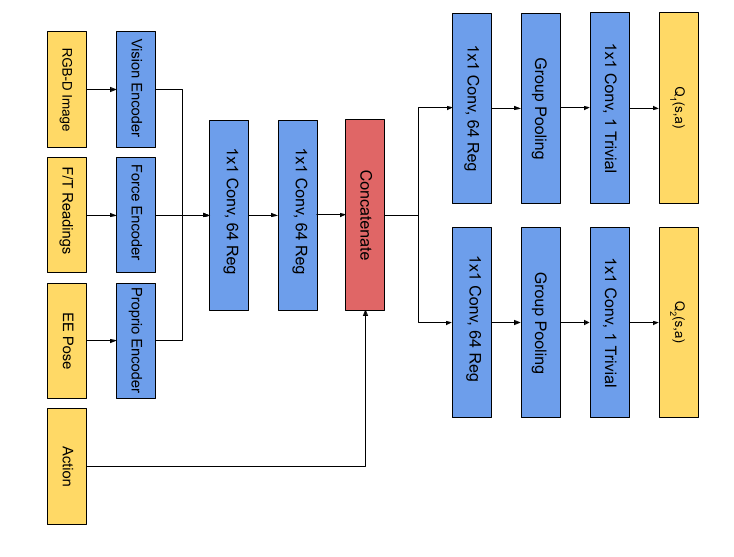}
        \caption{\textbf{Equivariant Critic Architecture.}}
    \end{subfigure}
    \caption{\textbf{Equivariant Soft Actor-Critic Architecture.} Network architectures of the equivariant actor and critic used in SVFL.}
    \label{fig:emf_sac}
\end{figure}

\begin{table}
    \centering
    \begin{tblr}{
    hline{1,2,3},
    columns = {halign=c}
    }
        Network & SVFL & CNN & VTT & POE \\
        \# of Parameters & 2.4E6 & 2.5E6  & 1.19E6 & 2.9E5 \\
    \end{tblr}
    \vspace{1em}
    \caption{Number of trainable parameters of SVFL, CNN, VTT, and POE in the reinforcement learning robotic manipulation tasks. Notice that due to being latent 
             representation learning methods, VTT and PoE utilize shared encoders between the actor and the critic so that the number of parameters is smaller than
             SVFL and CNN. Additionally, we utilize a smaller PoE as increasing the size of PoE has been shown to worsen performance\cite{chen2022visuo}.}
    \label{tab:emf_cnn_policy_params}
\end{table}

\subsubsection{Simulated Manipulation Policy Learning}

\noindent \textbf{SVFL/CNN Implementation Details:} Figure \ref{fig:encoder_model_architectures} shows the network architecture of the equivariant encoders 
and Figure \ref{fig:emf_sac} shows the network architecture of the Equivariant SAC in Section \ref{sec:experiments:policy}. The CNN network mimics the structure 
of the SVFL network but the equivariant convolutions are replaced with normal convolutions. In order to provide a fair comparison between the two methods, we 
increase the number of kernels in the CNN model such that the two methods have a comparable number of trainable parameters (Table \ref{tab:emf_cnn_policy_params}).
The equivariant network is implemented using the escnn \cite{cesa2022a,e2cnn} library, where all the hidden layers are defined using the regular representations. 
For the critic, the output is a trivial representation. For the actor, the output is a mixed representation containing one standard representation for the $(x,y)$ 
actions, one signed representation for the $\theta$ action, and seven trivial representations for the $(\lambda, z)$ actions alongside the standard deviations of 
all action components.

\noindent \textbf{VTT/PoE Implementation Details:} We utilize the same network architectures for PoE and VTT as used in the original VTT work\cite{chen2022visuo}
which can be found here: \url{https://github.com/yich7045/Visuo-Tactile-Transformers-for-Manipulation}.

\subsubsection{Supervised Block Centering}
\begin{figure}
    \begin{subfigure}[c]{0.45\textwidth}
        \includegraphics[width=\hsize]{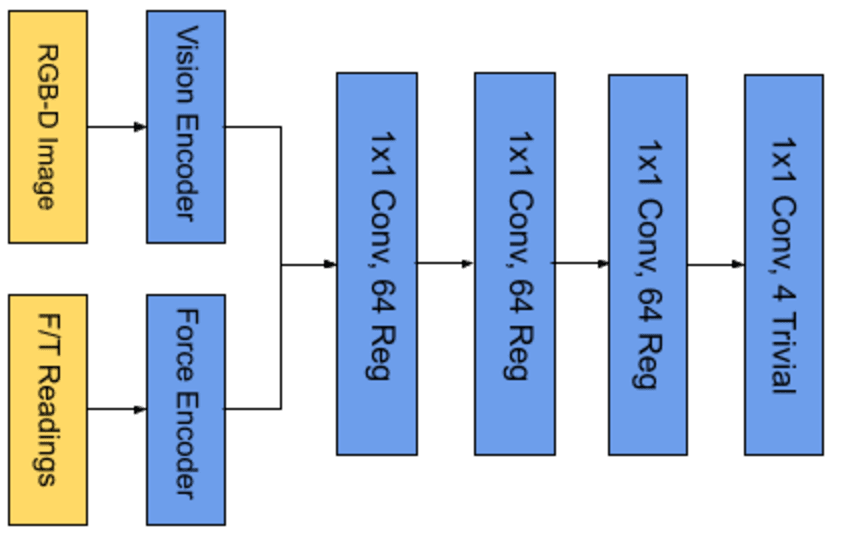}
    \end{subfigure}
    \hspace{5em}
    \begin{subfigure}[c]{0.45\textwidth}
        \begin{tblr}{
         hline{1,2,3},
         columns = {halign=c}
         }
             Network & SVFL & CNN  \\
             \# of Parameters & 2.3E6 & 2.2E6 \\
         \end{tblr}
    \end{subfigure}
    \caption{\textbf{Equivariant Classifier.} (Left) The network architecture used in the supervised learning experiment. (Right) Number of trainable 
             parameters of equivariant (SVFL) and conventional (CNN) models in the supervised learning task.}
    \label{fig:emf_classifier}
\end{figure}

We utilize the same network architectures for the vision and force encoders as in the reinforcement learning tasks (Figure \ref{fig:encoder_model_architectures}).
The two representations are combined using 2 convolutional layers and a final convolutional layer acts as the classification layer (Figure \ref{fig:emf_classifier}).
Figure \ref{fig:emf_classifier} (Right) shows the numbers of trainable parameters in both networks, where both networks have a similar number with SVFL having 
slightly more.

\subsection{Training Details}
\label{sec:appendix:training_details}

\subsubsection{Simulated Robotic Manipulation}
We utilize the manipulation tasks detailed in Section \ref{sec:appendix:tasks}. The workspace's size is $0.4m \times 0.4m \times 0.26m$. The minimum z height 
is slightly beneath the table allowing the arm to come in contact with the table. The pixel size of the visual observation is $4 \times 76 \times 76$ and is 
cropped to $4 \times 64 \times 64$ during training and testing. We utilize a random crop during training and a center crop during testing. The force data consists
of the most recent $64$ readings from the F/T sensor. We zero the force data using the first reading from the sensor after resetting the arm to its home position.
The maximum movement allowed for any action is limited to $\Delta x, \Delta y, \Delta z \in [-2.5cm, 2.5cm]$, $\Delta \theta \in [-\frac{\pi}{16}, \frac{\pi}{16}$, 
$\lambda \in [e_{min}, e_{max}]$ where $e_{min}$ and $e_{max}$ are the joint limits of the gripper. During training, we use 5 parallel environments where a training
step is performed after all 5 parallel environments perform an action step. The evaluation is performed every $500$ training steps. The training is implemented
in PyTorch \cite{paszke2017automatic}.

\noindent \textbf{SVFL/CNN Training:} We train using the Adam optimizer \citep{kingma2014adam} and the best learning rate and its decay were chosen to 
be $10^{-3}$ and $0.95$ respectively. The learning rate is multiplied by the decay every 500 training steps. We use the prioritized replay 
buffer \citep{schaul2015prioritized} with prioritized replay exponent $\alpha = 0.6$ and prioritized importance sampling exponent $B_0 = 0.4$
annealed to $1$ over training. We use a batch size of 64.

\noindent \textbf{VTT/PoE Training:} We pretrain the dynamics model for both VTT and PoE for $10,000$ steps as in \cite{chen2022visuo}. We train using 
the Adam optimizer \citep{kingma2014adam} using a learning rate of $10^{-4}$ for the latent model and a batch size of $32$. For policy training, we use a 
learning rate of $30^{-4}$ and a batch size of $64$. We use the prioritized replay  buffer \citep{schaul2015prioritized} with prioritized replay exponent
$\alpha = 0.6$ and prioritized importance sampling exponent $B_0 = 0.4$ annealed to $1$ over training.

\subsubsection{Block Centering}
The block is located in a workspace with a size of $0.4m \times 0.4m \times 0.26m$. The pixel size of the visual observation is $4 \times 76 \times 76$ and is 
cropped to $4 \times 64 \times 64$ during training and testing. We utilize a random crop during training and a center crop during testing. The force data consists 
of the most recent $64$ readings from the F/T sensor. We zero the force data using the first reading from the sensor after resetting the arm to its home position.
We train using the Adam optimizer \cite{kingma2014adam} with a learning rate of $10^{-3}$ and a cross-entropy loss. We use a batch size of $64$. 
The training is terminated when the test prediction success rate does not improve for $50$ epochs or when the maximum epoch of $500$ is reached.

\subsection{Additional Experiments}
\label{sec:appendix:additonal_exp}

\subsubsection{Effect of Dihedral Group Size}
\label{sec:appendix:dn_size}
\begin{figure}
    \centering
    \begin{subfigure}[c]{0.45\textwidth}
      \includegraphics[width=\textwidth]{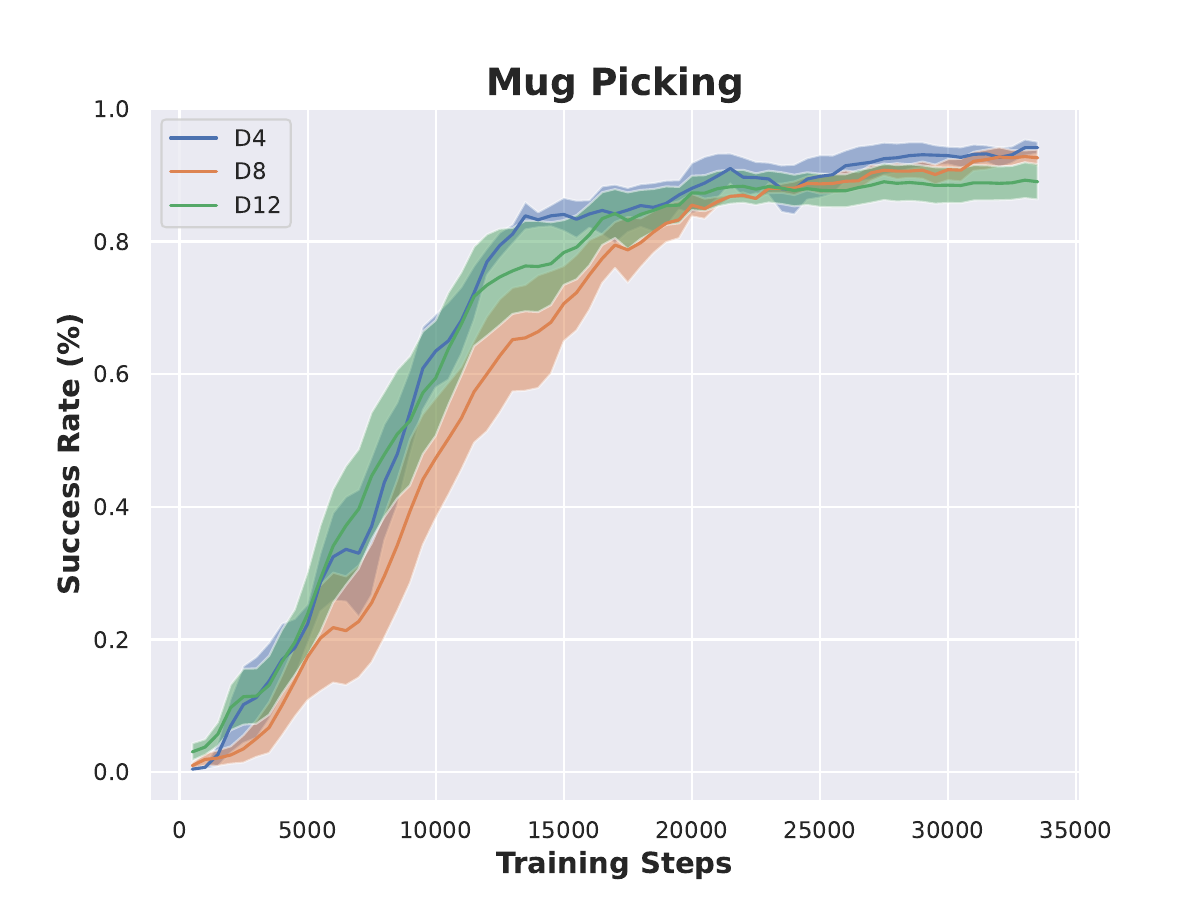}
    \end{subfigure}
    \begin{subfigure}[c]{0.45\textwidth}
      \includegraphics[width=\textwidth]{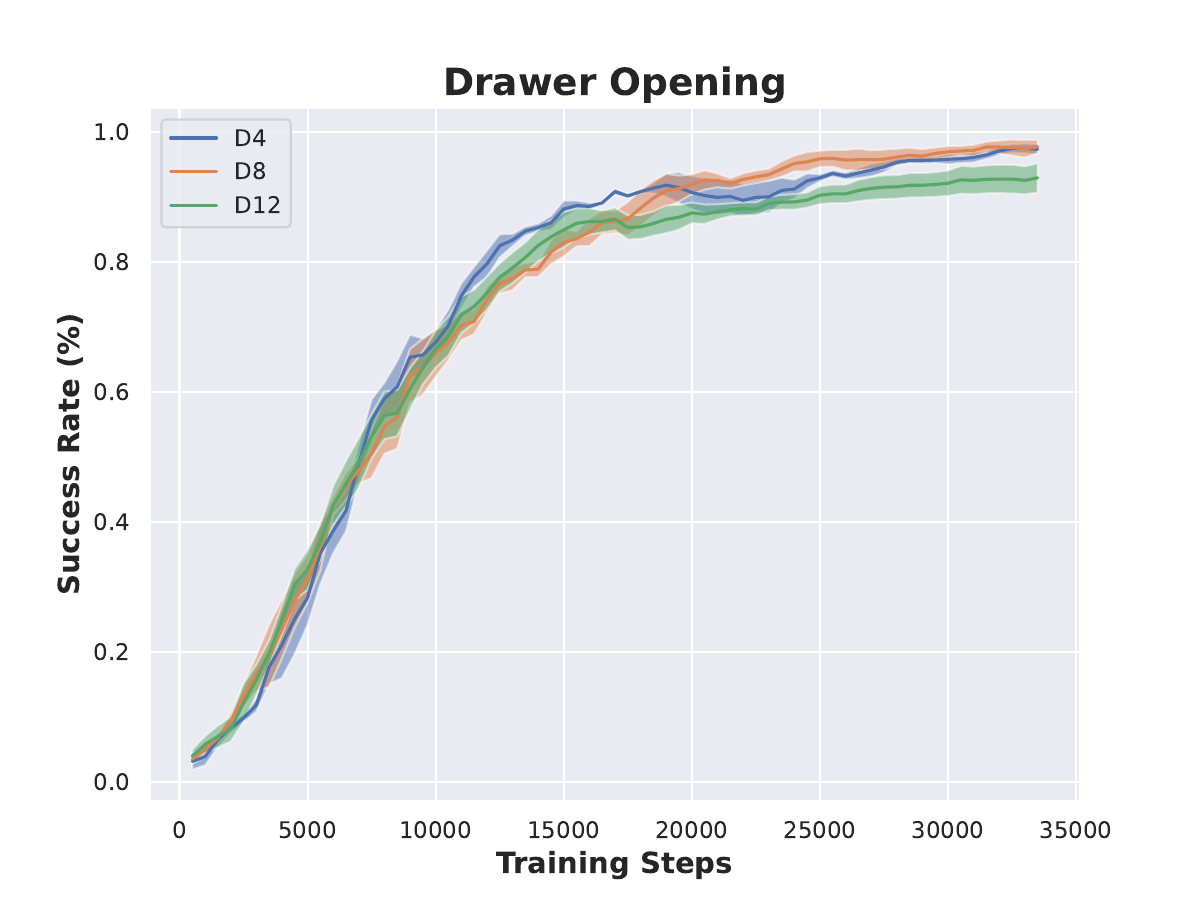}
    \end{subfigure}
    \caption{\textbf{Symmetry Group Size Comparison.} Effect of symmetry group on Mug Picking and Drawer Opening tasks.}
    \label{fig:dn_size}
\end{figure}

We compare the performance of SVFL in simulation while varying the size of the symmetry group. In this ablation, we are primarily interested in exploring the effect 
of the size of the discrete dihedral group as prior works have shown that this discrete group outperforms the continuous group $SO(2)$ in both supervised learning
\cite{weiler2019general} and reinforcement learning \cite{wang2022robot}. Figure \ref{fig:dn_size} shows the effect of increasing the size of the dihedral group 
for $d_4$, $d_8$, and $d_{12}$. While the overall performance is similar, we note an exponential increase in computational time as the size of the group
increases (Table \ref{tab:Dn_wall_time}). 

\begin{table}
    \centering
    \begin{tblr}{
    hline{1,2,3},
    columns = {halign=c}
    }
        Group Size   & 4   & 8   & 12   \\
        Time (Hours) & 5.4 & 7.3 & 16.7 \\
    \end{tblr}
    \vspace{1em}
    \caption{\textbf{Symmetry Group Wall-Clock Training Time.} Average wall-clock training time for the Mug Picking task for different symmetry group sizes.
             Averaged over three runs.}
    \label{tab:Dn_wall_time}
\end{table}

\subsubsection{Real World Block Centering}
\begin{figure}
    \centering
    \begin{subfigure}[c]{0.3\textwidth}
      \includegraphics[width=\textwidth]{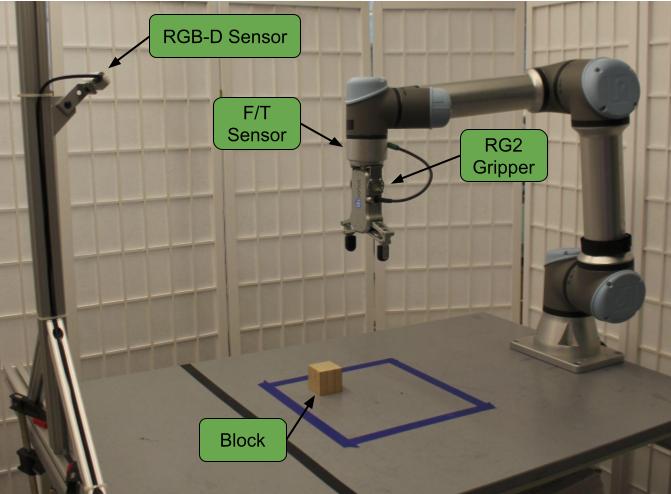}
    \end{subfigure}
    \begin{subfigure}[c]{0.69\textwidth}
    \small
    \begin{tblr}{
    hline{1,2,3,5,7,9},
    cell{1}{3} = {c = 4}{halign = c},
    cell{2}{1} = {c = 2}{halign = c},
    cell{3}{1} = {r = 2}{valign = m},
    cell{5}{1} = {r = 2}{valign = m},
    cell{7}{1} = {r = 2}{valign = m},
    columns = {halign=c}
    }
                &     & \# of Training Samples                                           \\
       Encoder  &     & $10$           & $25$           & $50$           & $100$          \\
       V        & CNN & $38.7 \pm 1.3$ & $41.4 \pm 2.5$ & $35.1 \pm 6.6$ & $59.5 \pm 5.8$ \\
                & SVFL & $51.3 \pm 3.4$ & $59.5 \pm 4.6$ & $67.7 \pm 3.8$ & $94.6 \pm 4.4$ \\
       F        & CNN & $21.6 \pm 1.3$ & $36.0 \pm 2.5$ & $42.3 \pm 1.3$ & $45.9 \pm 6.6$ \\
                & SVFL & $30.6 \pm 3.3$ & $54.5 \pm 2.2$ & $81.1 \pm 3.8$ & $92.8 \pm 1.3$ \\
       V+F      & CNN & $38.7 \pm 2.5$ & $38.7 \pm 5.5$ & $45.0 \pm 3.4$ & $73.9 \pm 2.5$ \\
                & SVFL & $57.8 \pm 4.4$ & $63.2 \pm 2.5$ & $87.6 \pm 9.2$ & $98.4 \pm 1.3$ \\
    \end{tblr}
    \end{subfigure}
    \caption{\textbf{Experiment on Robotic Hardware.} (Left) Robotic setup. (Right) Prediction accuracy (\%) on the test set for models trained with different
             amounts of training data. We compare the performance of equivariant and non-symmetric versions of the vision encoder (V), the force encoder (F), and the 
             fusion of these two encoders (V+F). Mean and standard error is given over three runs.}
    \label{fig:block_centering}
\end{figure}

We conduct an experiment using the Block Picking task to evaluate how well our model can leverage force observations from real-world robot
interactions. Here we do supervised learning rather than policy learning in order to focus on the model itself
and to reduce the variance of our results. Figure~\ref{fig:block_centering} shows the 
experimental setup which includes a UR5e robotic arm, a RG2 Gripper, a wrist-mounted force-torque sensor, and a Intel RealSense camera. The block
is a $5mm$ wooden cube that is randomly posed in the workspace. We learn a function, $h : (I, f) \mapsto \{0,1\}^4$, that maps visual-force 
observations to a four-way classification denoting the direction in which the gripper would need to move in order to grasp the block after a 
finger collides with the block. The idea was to mimic the most common failure case we see during policy learning
in block picking where the grasp was slightly offset from the block. In simulation, we observed that the force aware policy was able to determine
the correct direction to move to correct the failed grasp more often than the models without force.
The dataset is generated by a human teleoperator where each sample is the most recent sensor observations 
immediately following the collision. The goal of the teleoperator was to mimic a failed grasp where one finger came into contact with the block. 
We generate $200$ data samples and split the dataset into $100$ training samples and $100$ testing samples. We generated a diverse set of interactions 
between the block and the gripper varying the position of the gripper in relation to the block, the amount of force (by varying the amount of 
movement when coming into contact with the block), and the pose of the block.

We compared the classification accuracy of the baseline SVFL model against the non-symmetric version of the model with a similar number of trainable 
parameters (Section~\ref{sec:appendix:network_arch}, \ref{sec:appendix:training_details}). We examine the effect of three different types of input: 
Vision Only (V), Force Only (F), and Vision \& Force (V+F).  In each case, in order to measure the models' ability to generalize, we evaluated the 
performance on training sets of differing sizes including 10, 25, 50, and 100 samples. Figure~\ref{fig:block_centering} shows the accuracy of 
the models on the held-out test dataset. Notice that in all cases, the symmetric model does much better than its non-symmetric counterpart, both for differently 
sized training sets as well as for all input types.

\subsubsection{Simulated Block Centering}
\label{sec:appendix:sim_block_centering}
We repeat the real-world block centering supervised learning task in simulation using the BulletArm block picking domain. We generate the simulated block centering 
dataset by modifying the state-based planner such that one of the gripper fingers comes in contact with the block while attempting a picking action. Similar to the 
real-world dataset, we utilize a RGB-D sensor pointed at the center of the workspace and a wrist-mounted F/T sensor on a Franka Panda robot. We generate a dataset of 
$200$ samples and split this into $100$ training and $100$ test samples. We use the same SVFL and CNN modes as in Section \ref{sec:experiments:real_world_eval}. Table
\ref{tab:sim_block_centering} shows the test accuracy of both models on the held-out test dataset after being trained to convergence on training datasets of varying
sizes. Similar to the real-world results, SVFL outperforms the conventional encoders but the difference is much smaller in simulation, this is especially true when using
larger amounts of training data.

\begin{table}
    \centering
    \small
    \begin{tblr}{
    hline{1,2,3,5,7,9},
    cell{1}{3} = {c = 4}{halign = c},
    cell{2}{1} = {c = 2}{halign = c},
    cell{3}{1} = {r = 2}{valign = m},
    cell{5}{1} = {r = 2}{valign = m},
    cell{7}{1} = {r = 2}{valign = m},
    columns = {halign=c}
    }
                &     & \# of Training Samples                                            \\
       Encoder  &     & $10$           & $25$           & $50$           & $100$          \\
       V        & CNN & $48.5 \pm 1.8$ & $53.7 \pm 3.6$ & $72.2 \pm 2.4$ & $87.1 \pm 3.8$ \\
                & SVFL & $58.3 \pm 5.9$ & $66.2 \pm 3.4$ & $81.1 \pm 4.1$ & $98.6 \pm 2.3$ \\
       F        & CNN & $42.1 \pm 0.8$ & $48.8 \pm 1.6$ & $78.8 \pm 1.9$ & $87.2 \pm 3.2$ \\
                & SVFL & $48.4 \pm 1.7$ & $61.1 \pm 3.2$ & $89.2 \pm 2.3$ & $95.3 \pm 1.1$ \\
       V+F      & CNN & $56.8 \pm 1.4$ & $72.8 \pm 4.3$ & $92.1 \pm 6.2$ & $98.8 \pm 1.1$ \\
                & SVFL & $67.1 \pm 5.1$ & $77.3 \pm 2.2$ & $91.8 \pm 4.7$ & $99.3 \pm 0.4$ \\
    \end{tblr}
    \vspace{1em}
    \caption{\textbf{Simulated Block Centering.} Prediction accuracy (\%) on the test set for models trained with different amounts of simulated training data.
             We compare the performance of equivariant and conventional versions of the vision encoder (V), the force encoder (F), and the fusion of the two (V+F).
             Mean and standard error is given over three random seeds.}
    \label{tab:sim_block_centering}
\end{table}

\subsubsection{Additional Simulated Manipulation Tasks}
\begin{figure}
    \includegraphics[width=0.95\textwidth]{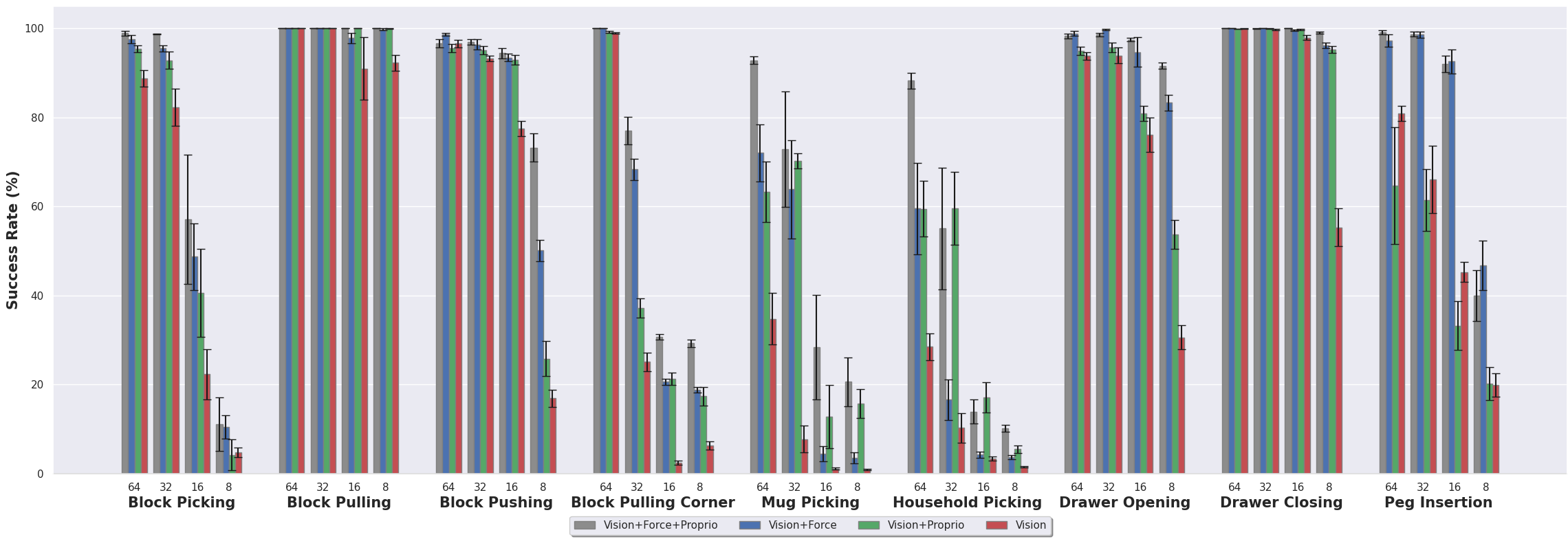}
    \caption{\textbf{Performance Under Degraded Visual Acuity.} Comparison of the full SVFL model (gray) versus SVFL with subsets of the data modalities under
             visual acuity degradation. Performance is given after all models are trained to convergence.}
    \label{fig:full_img_res_ablation}
\end{figure}

\begin{figure}
    \includegraphics[width=0.95\textwidth]{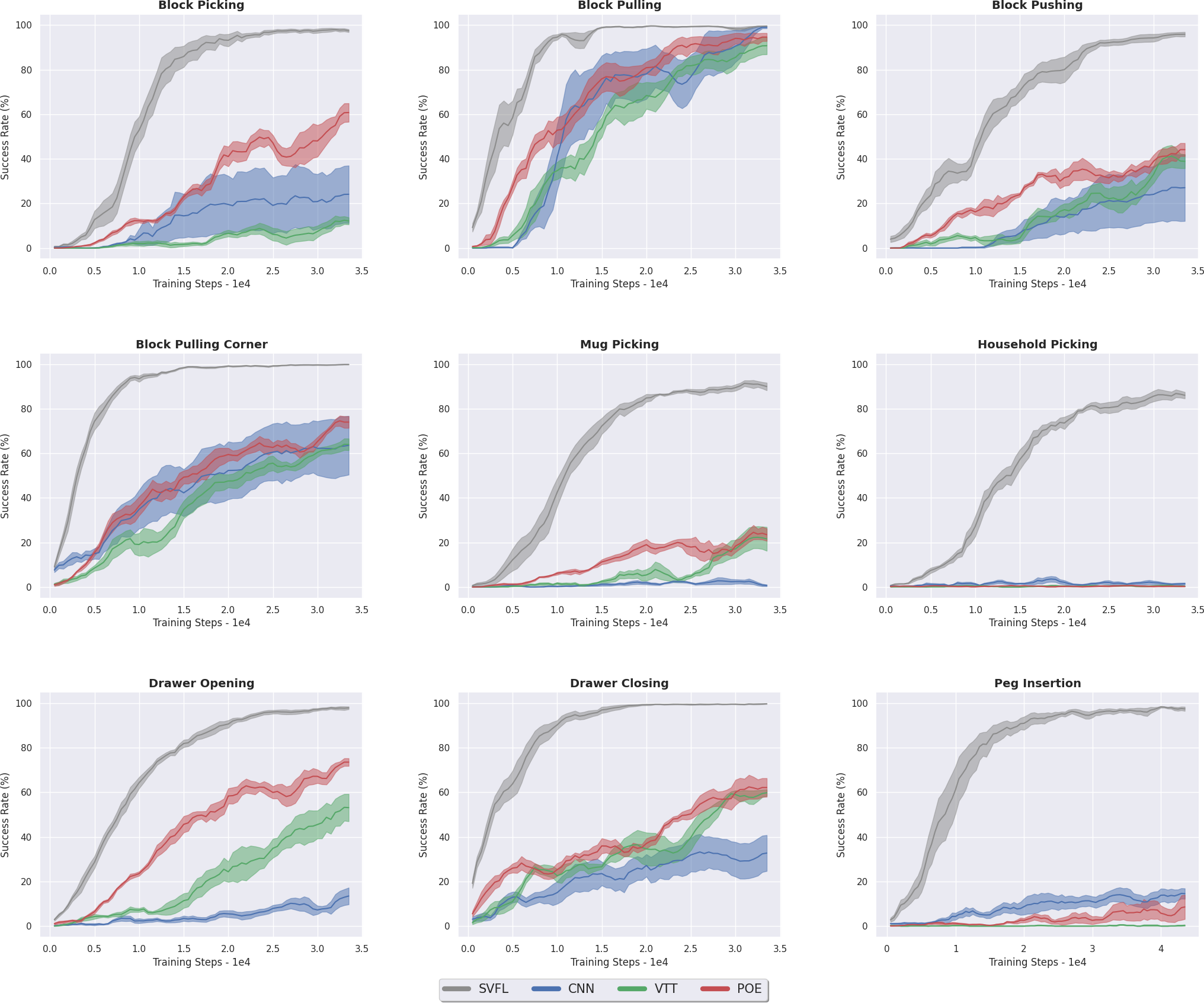}
    \caption{\textbf{Baseline Comparison.} Comparison of SVFL (gray) with baselines. Greedy evaluation policy is shown in terms of success rate. In all of 
             our experiments, results are averaged over 5 random seeds and the evaluation is performed every 500 training steps. Shading denotes standard error.}
    \label{fig:full_baseline_comparison}
\end{figure}

\begin{figure}
    \includegraphics[width=0.95\textwidth]{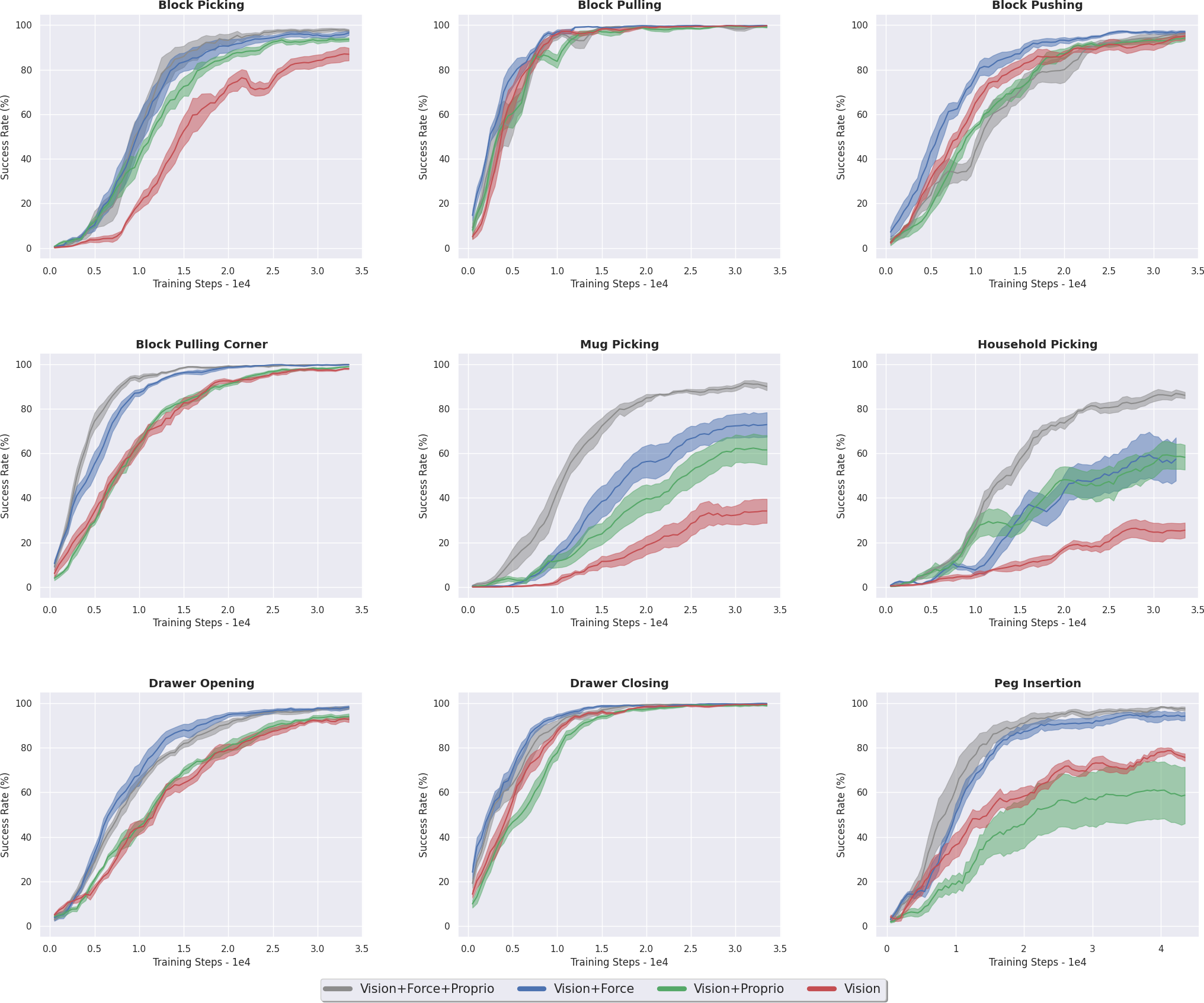}
    \caption{\textbf{Sensor Modality Ablation.} Comparison of the full SVFL model (gray) versus SVFL with subsets of the data modalities.}
    \label{fig:full_modality_ablation}
\end{figure}

In this section, we report the full results for the nine simulated manipulation tasks specified in Section \ref{sec:experiments:policy}. The baseline comparisons are 
shown in Figure \ref{fig:full_baseline_comparison}, the sensor modality ablations are shown in Figure \ref{fig:full_modality_ablation}, and the visual acuity ablations
are shown in Figure \ref{fig:full_img_res_ablation}.

\end{document}